\documentclass[preprint,12pt]{elsarticle}

\usepackage{graphicx}
\usepackage{subcaption} % cho (a), (b), (c)
\usepackage{float}      % để cố định vị trí hình
\usepackage{amssymb}
\usepackage{amsmath}
\usepackage{hyperref}
\usepackage{booktabs}
\usepackage{tabularx}
\hypersetup{
    colorlinks=true,
    linkcolor=blue,
    filecolor=magenta,      
    urlcolor=blue,
}
\journal{Nuclear Physics B}

\begin{document}

\begin{frontmatter}



\title{N-EIoU-YOLOv9: A Signal-Aware Bounding Box Regression Loss for Lightweight Mobile Detection of Rice Leaf Diseases} %% Article title

\author[1]{Dung Ta Nguyen Duc}
%\ead{dong.trinhcong@hust.edu.vn}

%\ead{Dung.TND240277E@sis.hust.edu.vn}
\author[1,2]{Thanh Bui Dang}
%\ead{thanh.buidang@hust.edu.vn}
\author[1]{Hoang Le Minh}
\author[1]{Tung Nguyen Viet\corref{cor1}}
%\ead{tung.nguyenviet@hust.edu.vn}
\author[1]{Huong Nguyen Thanh\corref{cor2}}
%\ead{huong.nguyenthanh3@hust.edu.vn}
\author[2]{Dong Trinh Cong\corref{cor3}}

\affiliation[1]{organization={School of Electrical and Electronic Engineering, Hanoi University of Science and Technology},
            addressline={No. 1, Dai Co Viet Road, Hai Ba Trung District}, 
            city={Hanoi},
            postcode={100000}, 
            state={},
            country={Vietnam}}
\affiliation[2]{organization={Institute of Control Engineering and Automation, Hanoi University of Science and Technology},
            addressline={No. 1, Dai Co Viet Road, Hai Ba Trung District}, 
            city={Hanoi},
            postcode={100000}, 
            state={},
            country={Vietnam}}
%\ead{Dung.TND240277E@sis.hust.edu.vn}
%\cortext[cor1]{Dung Ta Nguyen Duc}
\cortext[cor1]{Corresponding author: Tung Nguyen Viet, email: tung.nguyenviet@hust.edu.vn }
\cortext[cor2]{Corresponding author: Huong Nguyen Thanh, email: huong.nguyenthanh3@hust.edu.vn}
\cortext[cor3]{Corresponding author: Dong Trinh Cong, email: dong.trinhcong@hust.edu.vn}
%% Abstract
\begin{abstract}
Early and accurate detection of rice leaf diseases is critical for precision agriculture, yet deploying high-performance object detection models on resource-constrained mobile devices remains a significant challenge. In this work, we propose N-EIoU-YOLOv9, a lightweight detection framework centered on a signal-aware bounding box regression loss derived from non-monotonic gradient focusing and geometric decoupling principles, termed N-EIoU (Non-monotonic Efficient Intersection over Union). The proposed loss function explicitly reshapes localization gradients by integrating the non-monotonic focusing mechanism of N-IoU with the geometric decoupling properties of Efficient IoU (EIoU). From a signal processing perspective, N-EIoU enhances weak regression signals by amplifying gradients for hard samples with low overlap, while simultaneously reducing gradient interference between width and height optimization. This design is particularly effective for small-object and low-contrast targets, which are common in agricultural disease imagery. We integrate N-EIoU into the lightweight YOLOv9t architecture and evaluate the proposed method on a self-collected field dataset comprising 5,908 rice leaf images across four disease categories and healthy leaves. Experimental results demonstrate that N-EIoU consistently outperforms the standard CIoU loss, achieving a mean Average Precision (mAP@50) of 90.3\%, corresponding to a 4.3\% improvement over the baseline, with notable gains in mAP@50–95, indicating tighter and more accurate localization. To validate real-world applicability, the optimized model is deployed on an Android device using TensorFlow Lite with Float16 quantization, achieving an average inference time of 156 ms per frame while maintaining accuracy. These results confirm that the proposed N-EIoU provides an effective balance between localization accuracy, optimization stability, and computational efficiency by explicitly reshaping regression signals for hard samples, making it well suited for edge-based agricultural monitoring systems.
\end{abstract}

\begin{graphicalabstract}
\begin{figure}[H]
    \centering
    \includegraphics[width=1.0\textwidth]{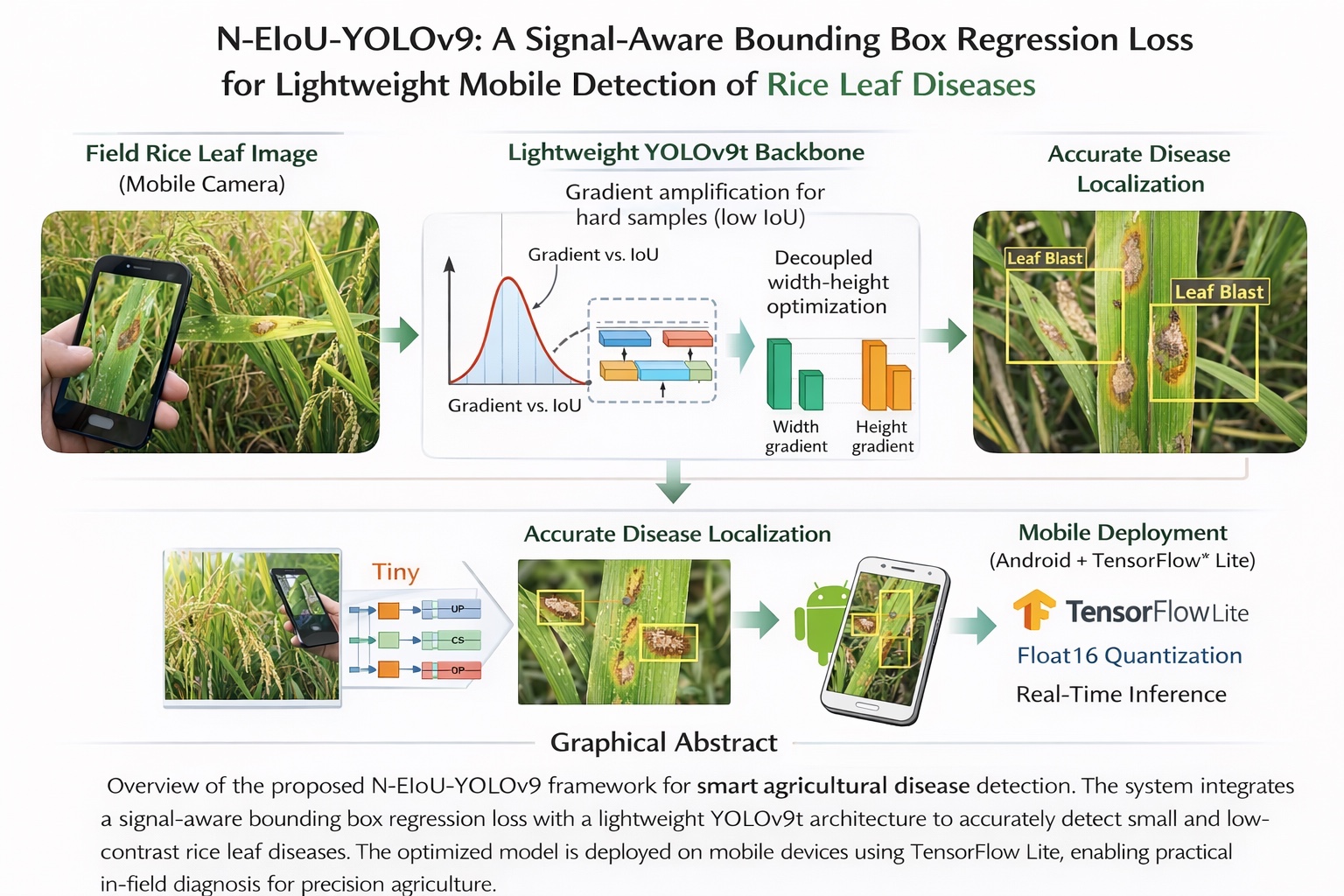}
    \caption{Graphical Abstract} 
    \label{fig:GraphAbs}
\end{figure}
\end{graphicalabstract}

\begin{highlights}
\item Proposal of an improved lightweight detection model based on YOLOv9t for rice leaf diseases.
\item Introduction of a novel \textbf{N-EIoU loss function}, integrating focal mechanism and geometric efficiency to resolve hard sample imbalance.
\item Superior performance achieving \textbf{90.3\% mAP@50}, resulting in a relative improvement of approximately 4\% in mAP@50, with more substantial gains observed under stricter IoU thresholds.
\item Successful implementation of \textbf{Float16 quantization} for model optimization.
\item Practical deployment on Android devices via TensorFlow Lite with an average inference time of 156ms.
\end{highlights}

%% Keywords
\begin{keyword}
Rice leaf diseases, YOLOv9, N-EIoU Loss, Computer Vision, TensorFlow Lite, Edge AI.


\end{keyword}

\end{frontmatter}



\section{Introduction}
\label{sec1}
The deployment of Convolutional Neural Networks (CNNs) on edge devices has revolutionized agricultural monitoring. However, detecting rice leaf diseases presents a unique "small object detection" challenge. Pathogens like Leaf Blast and Brown Spot often manifest as tiny necrotic lesions occupying less than 5\% of the image area, creating extreme visual ambiguity against background noise.
State-of-the-art models like YOLOv9 typically using CIoU loss for bounding box regression. While effective for general objects, we identify two critical limitations of CIoU in agricultural contexts. First, gradient vanishing for hard samples: as demonstrated by the N-IoU theory, standard IoU-based loss functions exhibit attenuated gradient responses when the overlap between the predicted and ground-truth bounding boxes is low. This causes the model to learn slowly on small, hard-to-hit lesion targets. Second, geometric coupling: the aspect ratio penalty term (v) in CIoU introduces coupled gradient responses for width (w) and height (h), leading to gradient interference during optimization. If w and h scale proportionally, the penalty term becomes zero, failing to drive the box to the correct size.

From a signal processing perspective, most existing IoU-based regression losses are designed primarily from geometric considerations, with limited attention to how gradient signals are distributed across samples of varying difficulty. In scenarios dominated by small, low-contrast objects, such as early-stage rice leaf diseases, the regression signal associated with hard samples is often severely attenuated, leading to suboptimal convergence. Despite recent advances in IoU variants, a unified signal-aware formulation that simultaneously amplifies weak localization signals and avoids geometric gradient interference remains underexplored.

To bridge this gap, we propose N-EIoU, a hybrid loss function designed to act as a gradient-shaping mechanism. By synthesizing the gradient-boosting capability of N-IoU and the dimension-decoupling of EIoU, our method prioritizes "hard samples" and ensures precise localization. This is integrated into a lightweight YOLOv9t architecture, optimized via Programmable Gradient Information (PGI) to prevent feature loss in deep layers. The main contributions of this work are threefold:

(1) We propose N-EIoU, a novel signal-aware bounding box regression loss that explicitly couples non-monotonic gradient focusing (via N-IoU) with geometric decoupling (via EIoU). Unlike existing IoU variants, N-EIoU reshapes regression gradients to prioritize hard, small-object samples while maintaining stable convergence.

(2) We provide a detailed analysis demonstrating that N-EIoU significantly improves localization accuracy under class imbalance and low-overlap conditions, yielding consistent gains in mAP@50–95 for small lesion targets commonly observed in agricultural imagery.

(3) We validate the practicality of the proposed loss through end-to-end deployment on mobile devices, showing that N-EIoU improves not only accuracy but also robustness under aggressive model compression (Float16 quantization) in edge environments.

\subsection{Related Works}
Nowadays, rapid advances in science and technology—especially in machine learning and deep learning—have opened up new opportunities to address challenges in agriculture. One of the most effective approaches is applying deep learning to plant images for disease detection and classification. In recent years, many studies have reported significant improvements in detection accuracy by using these techniques \cite{wijayanto2023machine, faisal2020gis}.

Deep learning aims to mimic how humans learn and make decisions, with convolutional neural networks (CNNs) playing a central role in tasks such as image recognition and pattern analysis \cite{krichen2023convolutional}. CNN-based models have been successfully applied in various fields, including healthcare, finance, education, and agriculture \cite{alahmari2023learning}. In the agricultural domain, deep learning has shown promising results for plant disease detection. For instance, \cite{saleem2019plant} studied foliar disease surveillance across different infection stages and analyzed the impact of factors such as lighting conditions, dataset size, and learning rate on model performance.

In addition to classification, object detection models have been widely adopted for leaf disease detection. YOLO-based approaches, ResNet architectures, and dedicated CNN models have been applied to detect and classify rice and tomato leaf diseases, achieving encouraging results \cite{agbulos2021identification, zhang2018can, liu2020tomato, sharma2022rice}. Several studies compared and combined multiple CNN architectures to improve accuracy, including DenseNet, ResNet variants, and SE-based models \cite{deng2021automatic}. High classification performance has also been reported using hybrid methods such as CNN–SVM and optimization-based deep learning techniques \cite{shrivastava2019rice, hasan2019rice, ramesh2020recognition}.

More recent works have focused on improving robustness under real-world conditions through advanced preprocessing, segmentation, and modern object detection models. Approaches using hue-based segmentation and YOLOv5 have demonstrated strong performance in detecting common rice leaf diseases such as blast, brown spot, bacterial blight, and sheath rot \cite{maheswaran2022detection, haque2022rice, jhatial2022deep}. Overall, these studies highlight the strong potential of deep learning as a practical and reliable solution for plant disease detection in agriculture.

\subsection{Evolution of Bounding Box Regression Losses}
\label{sec:related_work_losses}
Bounding Box Regression (BBR) is the cornerstone of localization in modern object detectors. The evolution of loss functions for BBR has progressed from naive geometric distances to complex metric-based penalties, aiming to bridge the gap between training objectives and evaluation metrics.

To clearly position the proposed N-EIoU loss with respect to existing IoU-based regression losses, we provide a qualitative comparison from a signal processing perspective in Table \ref{tab:iou_loss_comparison}. Unlike conventional geometric comparisons, this analysis emphasizes gradient behavior, signal strength for hard samples, and the degree of gradient coupling between bounding box dimensions. Such a comparison highlights how different loss formulations distribute regression signals during optimization. As shown, N-EIoU uniquely combines non-monotonic gradient emphasis with geometric decoupling, resulting in a signal-aware design tailored for small-object detection.

\begin{table}[H]
\centering
\caption{Qualitative comparison of IoU-based bounding box regression losses from a signal processing perspective}
\label{tab:iou_loss_comparison}
\begin{tabularx}{\linewidth}{l X X X X}
\toprule
\textbf{Loss Function} & \textbf{Gradient behavior} & \textbf{Performance on small objects} & \textbf{ w--h Gradient coupling} & \textbf{Signal-aware design} \\
\midrule
IoU   & Linear gradient attenuation & Limited, weak gradients for low-overlap samples & Strongly coupled & No \\
CIoU  & Linear gradient attenuation & Limited, similar to IoU & Strongly coupled via aspect ratio penalty & No \\
EIoU  & Linear gradient attenuation & Moderate improvement due to geometric decoupling & Decoupled & No \\
N-IoU & Non-monotonic gradient emphasis & Effective for hard samples with low overlap & Coupled & Partially \\
\textbf{N-EIoU(Ours)} & \textbf{Non-monotonic gradient emphasis with geometric decoupling} & \textbf{Highly effective for small and low-overlap objects} & \textbf{Decoupled} & \textbf{Yes} \\
\bottomrule
\end{tabularx}
\end{table}

\subsubsection{From $L_n$-norm to Metric-based Losses}
Early detectors like R-CNN and Fast R-CNN employed $L_n$-norm losses (e.g., Smooth $L_1$) to optimize the offset between prediction and ground truth coordinates. However, these losses treat the four coordinates $(x, y, w, h)$ as independent variables, ignoring the inherent geometric correlations of the bounding box. To address this, the Intersection over Union (IoU) loss was introduced, optimizing the box as a unified entity and achieving scale invariance.
Despite its theoretical elegance, standard IoU suffers from the ``gradient vanishing'' problem when boxes do not overlap ($IoU=0$). Subsequent variants like Generalized IoU (GIoU) and Distance IoU (DIoU) introduced penalty terms based on the smallest enclosing box and center point distance, respectively, to maintain gradient flow.

\subsubsection{Emerging Trends: Signal-Aware Gradient Shaping}
Recent advancements have shifted focus from purely geometric constraints to \textit{gradient shaping mechanisms}. The core idea is to dynamically re-weight the regression signals based on sample difficulty.
\begin{itemize}
    \item \textbf{Focal Loss intuition:} Inspired by Focal Loss in classification, methods like Focal-EIoU attempt to assign larger gradients to high-quality anchors to refine localization accuracy.
    \item \textbf{N-IoU and Non-monotonic Focusing:} More recently \cite{} introduced N-IoU, utilizing the Dice coefficient logic to reshape the gradient curve. Unlike standard IoU which provides linear gradients, N-IoU generates a non-monotonic curve that significantly amplifies gradients for samples with low overlap ($0.1 < IoU < 0.4$). This property is particularly crucial for detecting small disease lesions (hard samples) where standard losses provide weak supervision signals.
\end{itemize}

In this work, we synthesize the geometric decoupling of EIoU and the signal amplification of N-IoU into a unified framework, termed \textbf{N-EIoU}, to specifically address the localization challenges in rice leaf disease detection.

\subsection{Rice Common Diseases}
\begin{figure}[H] 
    \centering
    \begin{subfigure}{0.16\textwidth}
        \centering
        \includegraphics[width=\linewidth]{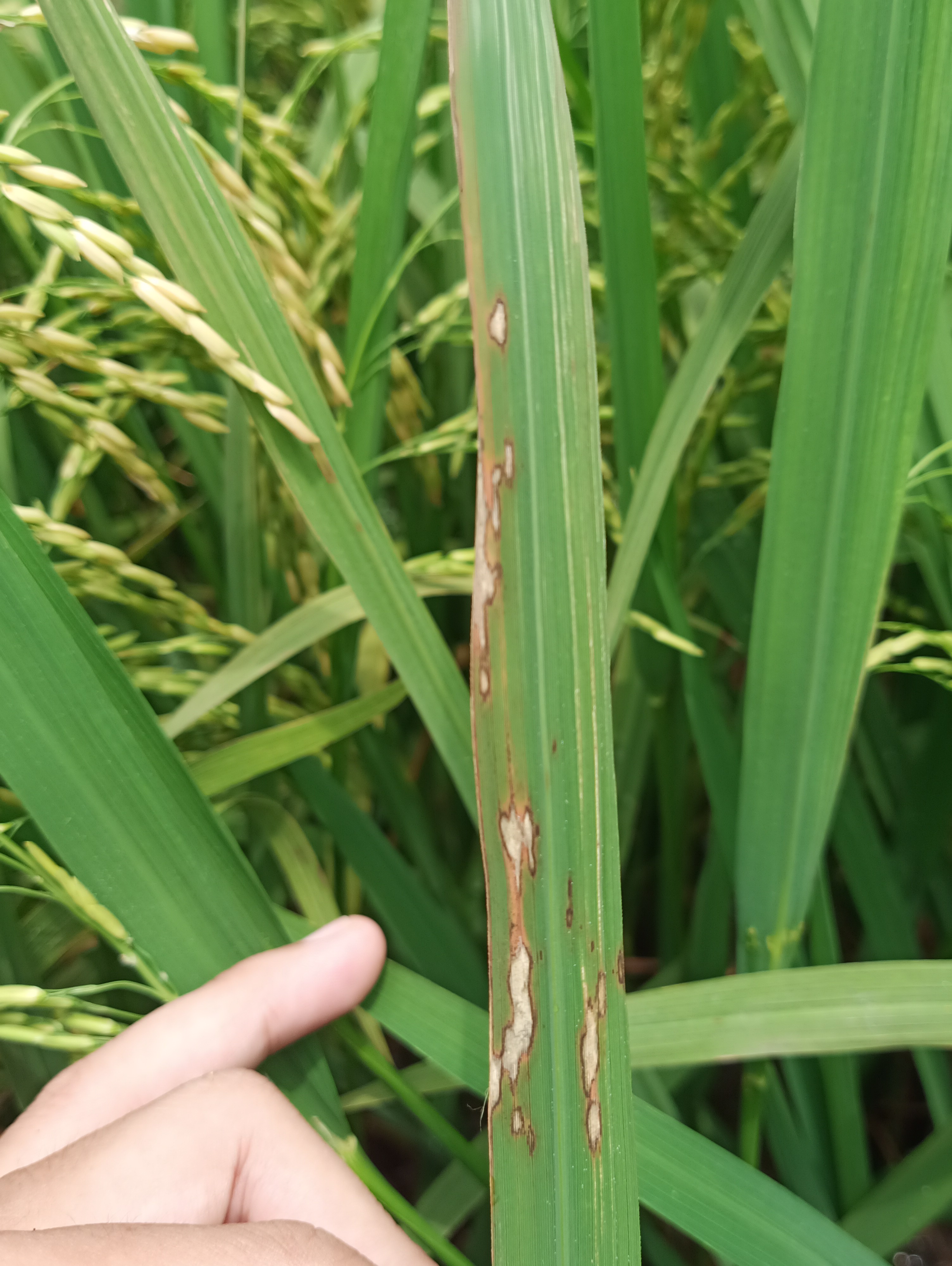}
        \caption{Leaf blast}
    \end{subfigure}
    \hfill
    \begin{subfigure}{0.16\textwidth}
        \centering
        \includegraphics[width=\linewidth]{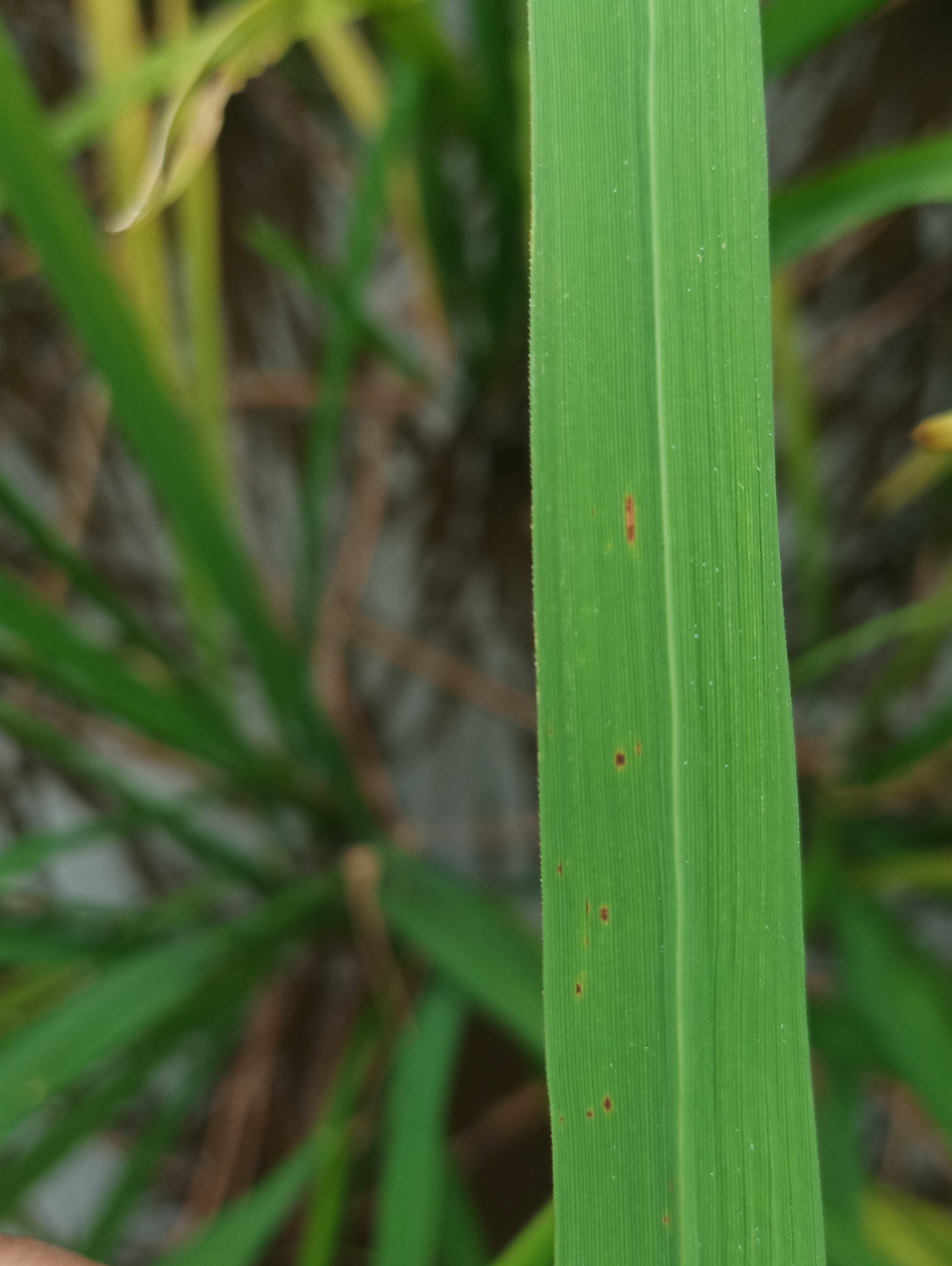}
        \caption{Brown spot}
    \end{subfigure}
    \hfill
    \begin{subfigure}{0.16\textwidth}
        \centering
        \includegraphics[angle=90,width=\linewidth]{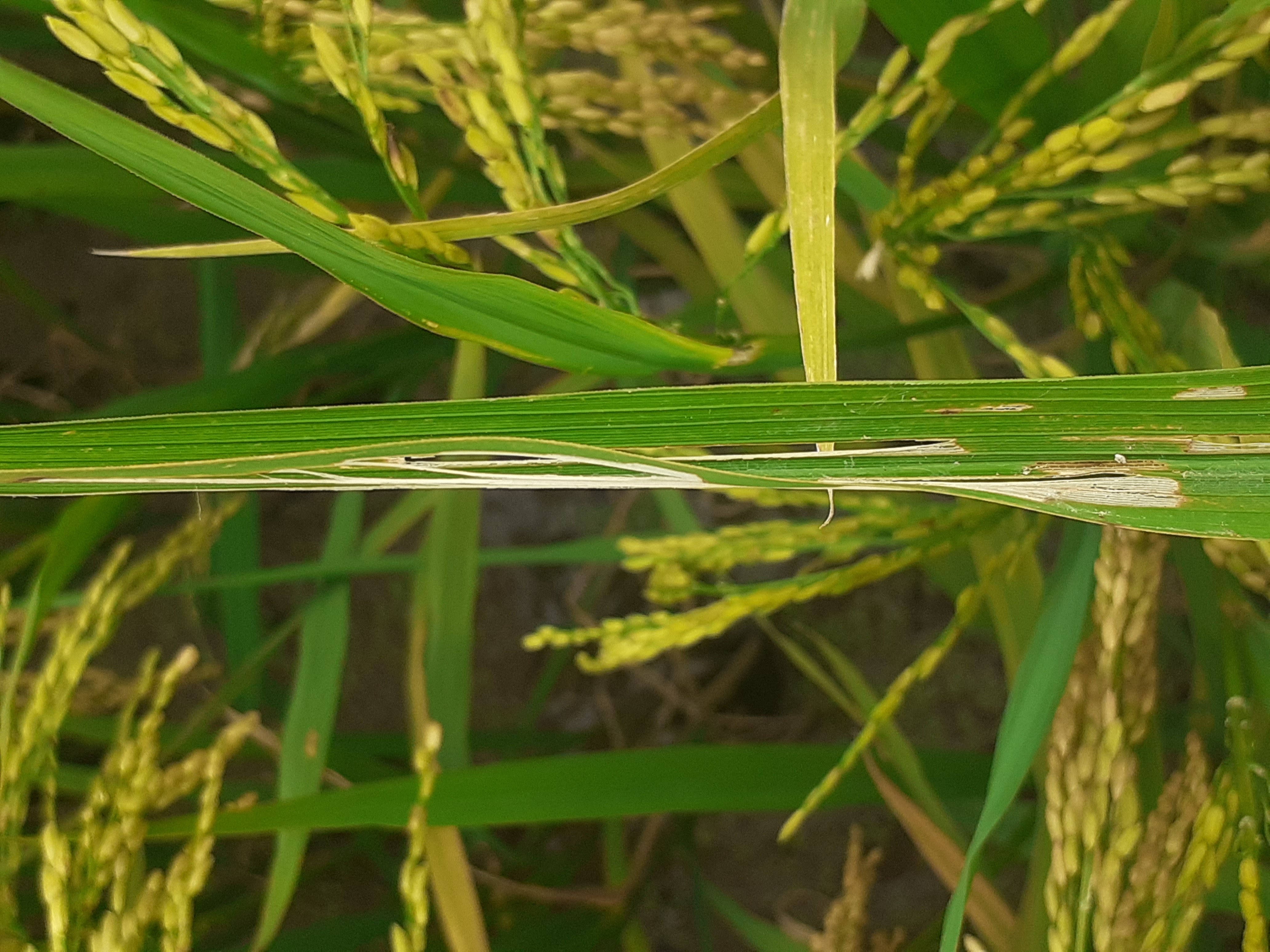}
        \caption{Leaf folder}
    \end{subfigure}
    \hfill
    \begin{subfigure}{0.16\textwidth}
        \centering
        \includegraphics[width=\linewidth]{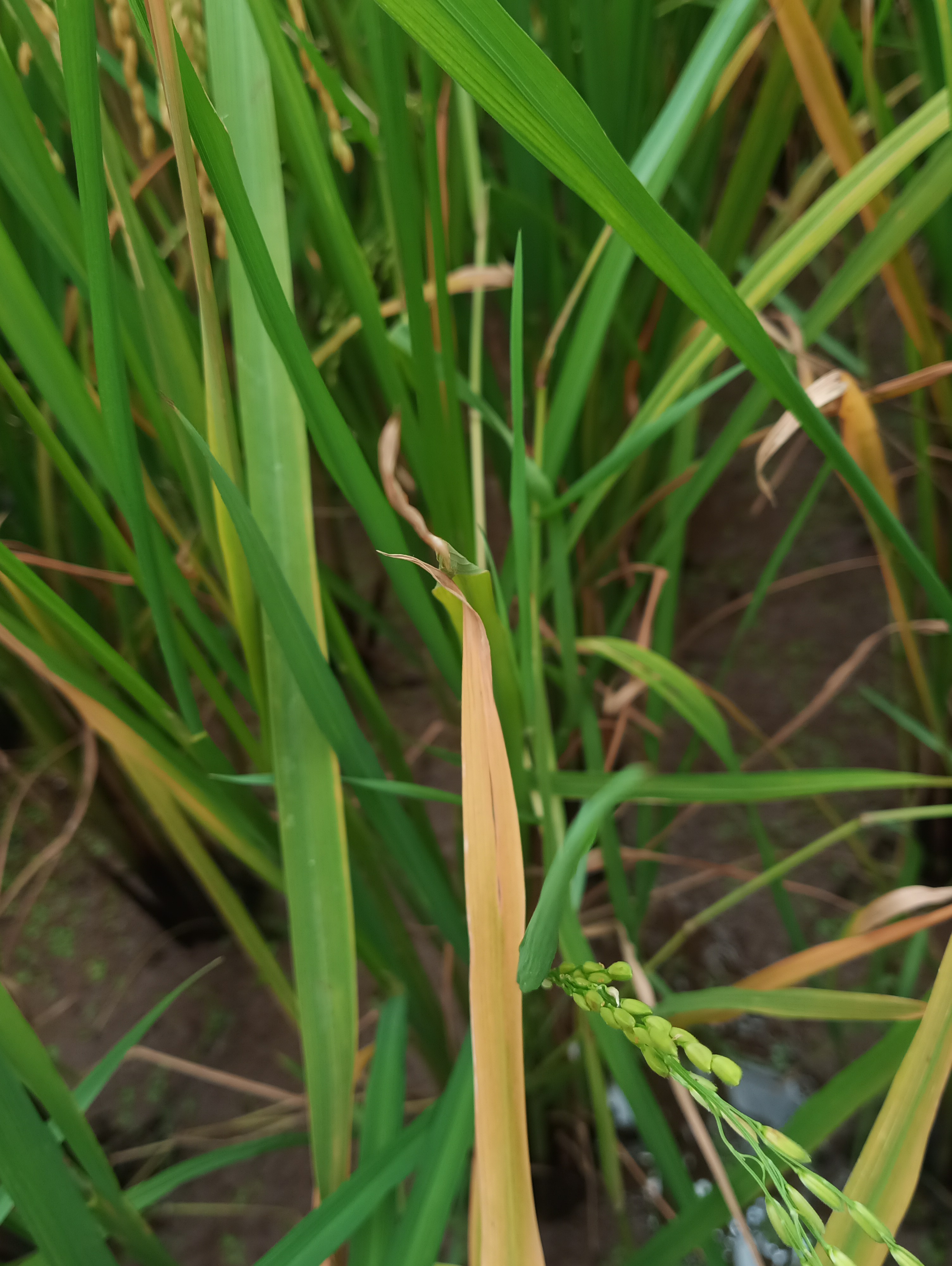}
        \caption{Yellow~dwarf}
    \end{subfigure}
    \hfill
    \begin{subfigure}{0.16\textwidth}
        \centering
        \includegraphics[width=\linewidth]{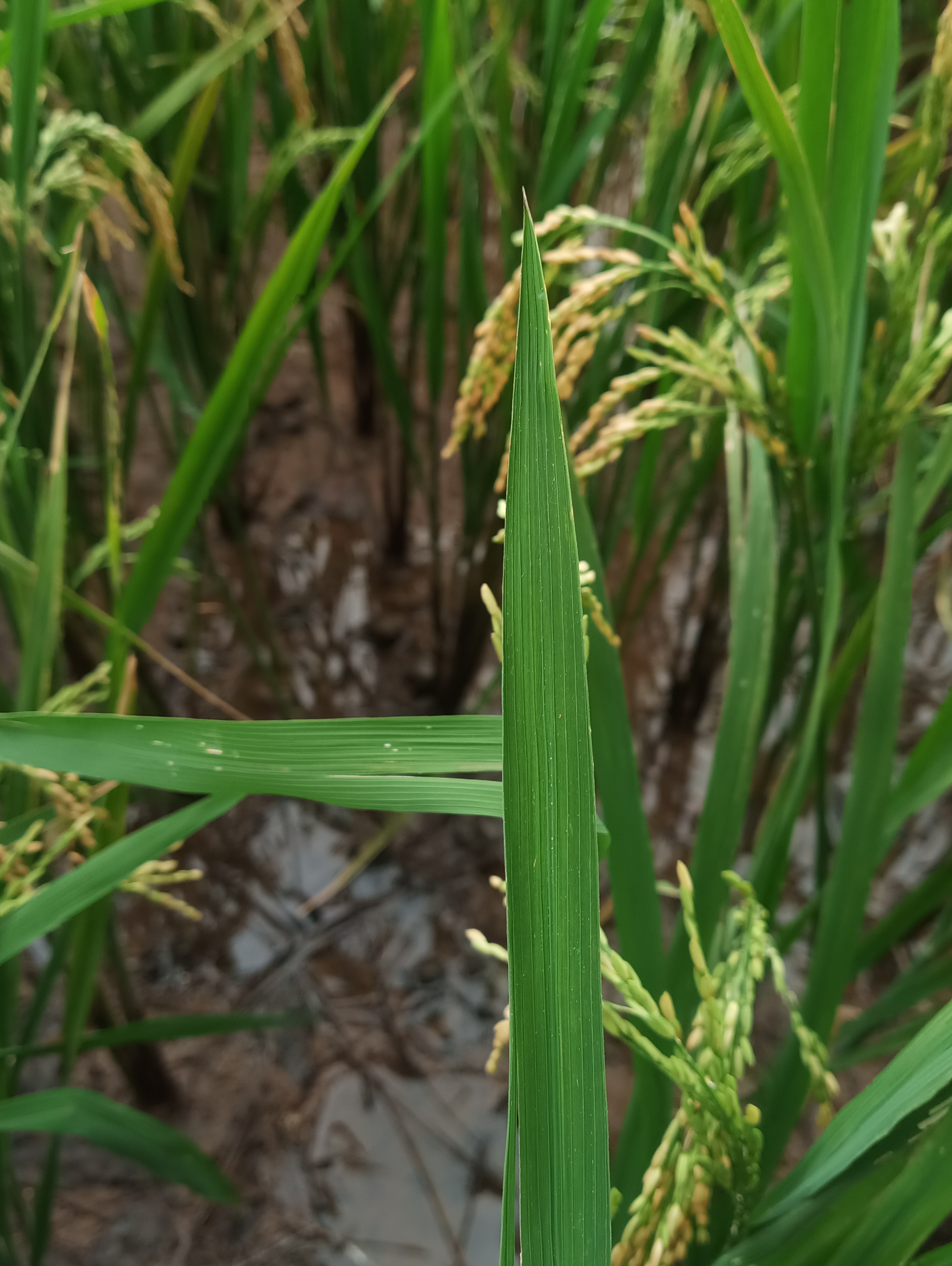}
        \caption{Healthy}
    \end{subfigure}    
    \caption{Common rice leaf diseases (a, b, c, d) in Vietnam.}
    \label{fig:rice_disease}
\end{figure}

Instead of delving into the biological pathology of pathogens like Magnaporthe oryzae or Bipolaris oryzae, this study focuses on the specific visual characteristics that impede automated detection performance. We target four prevalent rice diseases in Vietnam: \textbf{Leaf Blast}, \textbf{Brown Spot}, \textbf{Leaf Folder}, and \textbf{Yellow Dwarf}, along with \textbf{Healthy Leaves}. From a signal processing and object detection perspective, these categories present two distinct levels of difficulty:

• \textbf{Small-Object and Low-Contrast Targets (Hard Samples)}: Leaf Blast and Brown Spot typically manifest as tiny, necrotic lesions (often occupying less than 5\% of the image area). These lesions exhibit high visual similarity to background noise such as soil splashes, debris, or insect bites, leading to severe visual ambiguity. Standard loss functions (e.g., CIoU) often struggle with these "hard samples" due to weak supervision signals and the vanishing gradient problem when the Intersection over Union (IoU) is low.

• \textbf{Morphological and Chromatic Deformations}: Leaf Folder is characterized by longitudinal whitish streaks and physical folding of the leaf blade (morphological change), while Yellow Dwarf presents as a distinct discoloration of the entire leaf (chromatic change). While these features are more salient, they require the model to be robust against varying lighting conditions and geometric distortions.
The coexistence of these multi-scale targets—ranging from microscopic spots to whole-leaf discoloration—creates a significant challenge in bounding box regression, necessitating a loss function capable of gradient re-weighting to prioritize the difficult, small-scale lesions.

\section{Methodology}
We constructed the \textbf{DUNG\_BK65\_RiceLeafDiseases} dataset, comprising 5,908 images collected directly from fields of Vietnam National University of Agriculture (VNUA). All images in the proposed dataset were directly reviewed and verified by experts from the Vietnam National University of Agriculture.
The dataset includes four disease classes (\textit{Leaf Blast, Brown Spot, Leaf Folder, Yellow Dwarf}) and \textit{Healthy Leaves}. All the dataset 
From a computer vision perspective, these categories present two distinct levels of difficulty:
\begin{itemize}
    \item \textbf{Small-Object and Low-Contrast Targets (Hard Samples):} \textit{Leaf Blast} and \textit{Brown Spot} typically manifest as tiny, necrotic lesions with high visual similarity to background noise. Standard loss functions often struggle with these ``hard samples'' due to weak supervision signals.
    \item \textbf{Morphological and Chromatic Deformations:} \textit{Leaf Folder} and \textit{Yellow Dwarf} involve significant shape and color changes.
\end{itemize}

\subsubsection{Limitations of the State-of-the-Art CIoU}
State-of-the-art object detectors, including the baseline YOLOv9, typically employ CIoU (Complete IoU) loss for bounding box regression. The CIoU loss is defined as:
\begin{equation}
    \mathcal{L}_{CIoU} = 1 - IoU + \frac{\rho^2(\mathbf{b}, \mathbf{b}^{gt})}{c^2} + \alpha v
\end{equation}
where $\rho$ denotes the Euclidean distance between center points, $c$ is the diagonal length of the smallest enclosing box, and $\alpha v$ is the aspect ratio penalty. While CIoU accounts for overlap, distance, and aspect ratio, we identify three critical limitations when applied to the detection of small agricultural lesions:

\textbf{1. Ambiguity in Aspect Ratio Penalty:} The term $v$ measures the consistency of aspect ratios rather than the actual deviation in dimensions. As noted by Zhang et al. [EIoU Paper], if the predicted box width and height scale proportionally to the ground truth (i.e., $\{w = k \cdot w^{gt}, h = k \cdot h^{gt} | k \neq 1\}$), the penalty term $v$ becomes zero despite the existence of localization errors. This renders the loss function insensitive to scale discrepancies, which is detrimental for estimating the precise size of variable lesions like \textit{Leaf Blast}.

\textbf{2. Gradient Conflict (The "See-Saw" Effect):} The gradients of $v$ with respect to width ($w$) and height ($h$) are inversely related: $\frac{\partial v}{\partial w} = -\frac{h}{w} \frac{\partial v}{\partial h}$. This implies that during backpropagation, if the gradient updates urge the width to increase, they simultaneously force the height to decrease. This conflicting optimization creates a "see-saw" effect, hindering convergence speed when both dimensions need to be adjusted simultaneously to fit a target.

\textbf{3. Weak Supervision for Hard Samples:} From a signal processing perspective, CIoU inherits the linear gradient behavior of standard IoU. As analyzed in N-IoU theory [Su et al.], for "hard samples" (small lesions with low IoU overlap, e.g., $<0.3$), the magnitude of the gradient signal is relatively weak. In our rice disease dataset, where small \textit{Brown Spot} lesions dominate, this leads to the "gradient vanishing" problem, causing the model to prioritize easier, larger targets and neglect the difficult ones.

\subsection{The Proposed N-EIoU Loss Function}
We formulate the Bounding Box Regression (BBR) as a signal optimization problem. The proposed \textbf{N-EIoU Loss} combines the focusing mechanism of N-IoU with the geometric efficiency of EIoU:

\begin{equation}
    \mathcal{L}_{N\text{-}EIoU} = \mathcal{L}_{N\text{-}IoU} + \mathcal{L}_{dis} + \mathcal{L}_{asp}
\end{equation}

\subsubsection{Signal Amplification via N-IoU}
To address the gradient vanishing problem for small lesions (low IoU), we replace the standard IoU term with N-IoU \cite{}:
\begin{equation}
    \mathcal{L}_{N\text{-}IoU} = 1 - \frac{I + nI}{U + nI}
\end{equation}
where $I$ is the intersection area, $U$ is the union area, and $n$ is a hyperparameter (set to $n=9$). Unlike standard IoU which is linear, N-IoU creates a non-monotonic gradient curve. It provides \textbf{high gradient gain} when IoU is low ($0.1 - 0.4$), effectively amplifying the learning signal for small, misaligned boxes (hard samples).

\subsubsection{Geometric Decoupling via EIoU}
To eliminate the conflicting gradients in CIoU's aspect ratio penalty, we adopt the separated dimension penalties from EIoU \cite{}:
\begin{equation}
    \mathcal{L}_{dis} = \frac{\rho^2(b, b^{gt})}{c^2}, \quad \mathcal{L}_{asp} = \frac{\rho^2(w, w^{gt})}{C_w^2} + \frac{\rho^2(h, h^{gt})}{C_h^2}
\end{equation}
Here, we minimize $(w - w^{gt})^2$ and $(h - h^{gt})^2$ independently. This allows the model to refine the bounding box boundaries precisely without the ``see-saw'' effect observed in CIoU.

\subsubsection{Theoretical Interpretation}

From a signal processing perspective, N-EIoU can be interpreted as a \textbf{gradient-shaping mechanism}. The N-IoU component dynamically amplifies regression gradients for low-overlap samples, effectively acting as a non-linear emphasis filter for weak localization signals. Meanwhile, the EIoU term decouples width and height errors, reducing gradient interference during optimization. This combination yields a more stable and discriminative regression signal, which is particularly beneficial for small-object detection under noisy backgrounds.

\subsubsection{N-EIoU as a Gradient Signal Filtering Mechanism}
From a signal processing perspective, the mapping between Intersection over Union (IoU) and gradient magnitude can be viewed as a signal filtering mechanism that controls the distribution of regression energy across samples of varying difficulty. In conventional IoU-based losses such as CIoU, gradient magnitude decreases monotonically as IoU becomes smaller, causing hard samples with low overlap to receive weak corrective signals during optimization.

In contrast, N-IoU introduces a non-monotonic gradient emphasis that selectively amplifies regression signals in the low-IoU regime. Building upon this idea, N-EIoU further integrates orthogonal decomposition of width and height errors, enabling strong gradient responses for hard samples while mitigating cross-dimensional gradient interference. This signal-aware formulation provides a clear explanation for the improved convergence stability and localization accuracy observed in subsequent experiments.

\subsection{Lightweight YOLOv9t for Edge Inference}
We select \textbf{YOLOv9t (Tiny)} for this study. With approximately 2 million parameters, YOLOv9t represents the optimal trade-off between inference latency and detection performance. To mitigate the "information bottleneck" inherent in lightweight models, YOLOv9 employs \textbf{Programmable Gradient Information (PGI)} and \textbf{GELAN} blocks. PGI provides reliable gradient information to update weights across all layers, ensuring that deep features retain critical pixel-level details of small lesions (like Rice Blast) which are often lost in standard CNNs.
\begin{figure}[H]
    \centering
    \includegraphics[width=1.1\textwidth]{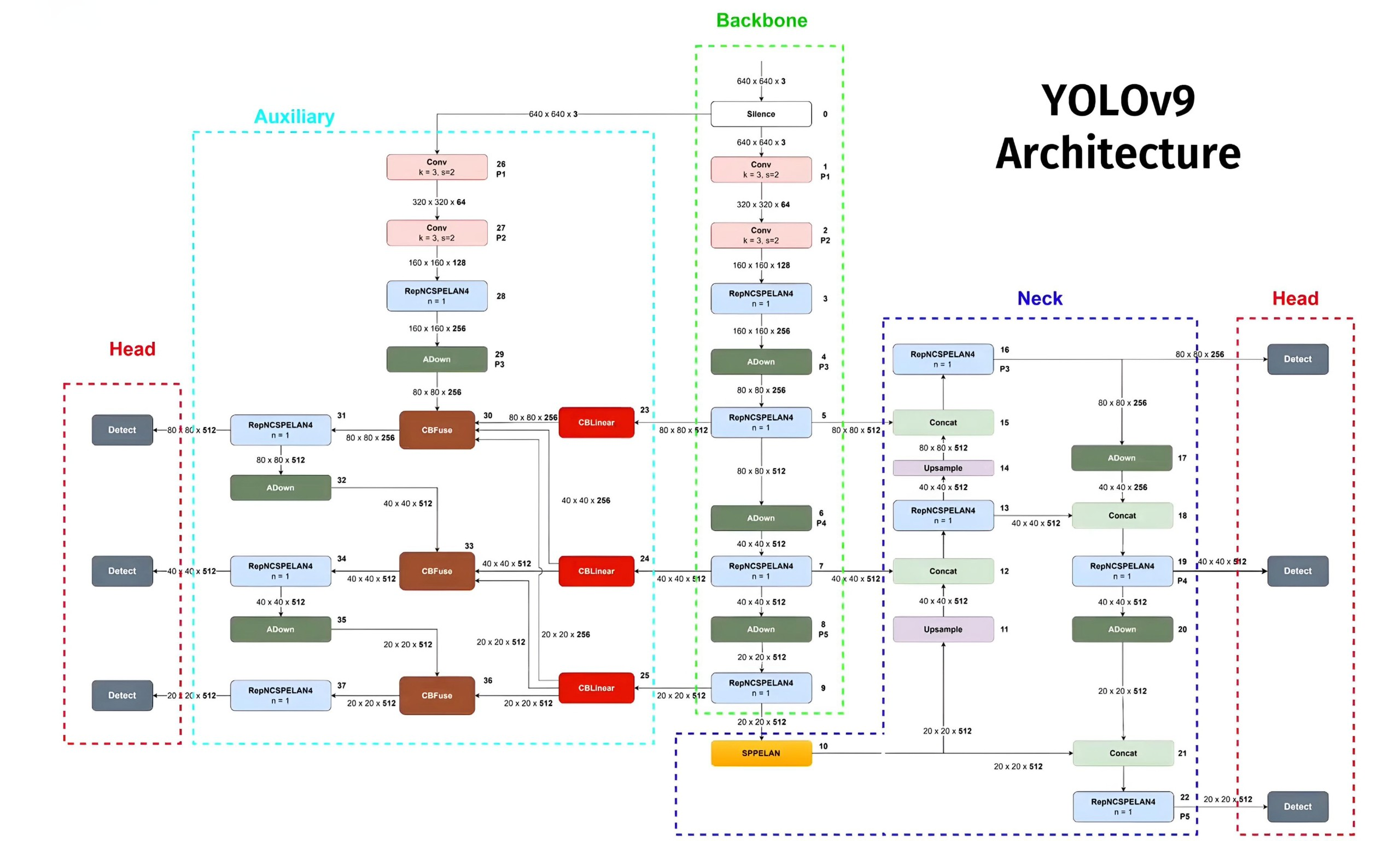}
    \caption{YOLOv9 structure} 
    \label{fig:simulation}
\end{figure}

\section{Experiments and Analysis}

\subsection{Gradient Simulation}
\begin{figure}[H]
    \centering
    \includegraphics[width=0.8\textwidth]{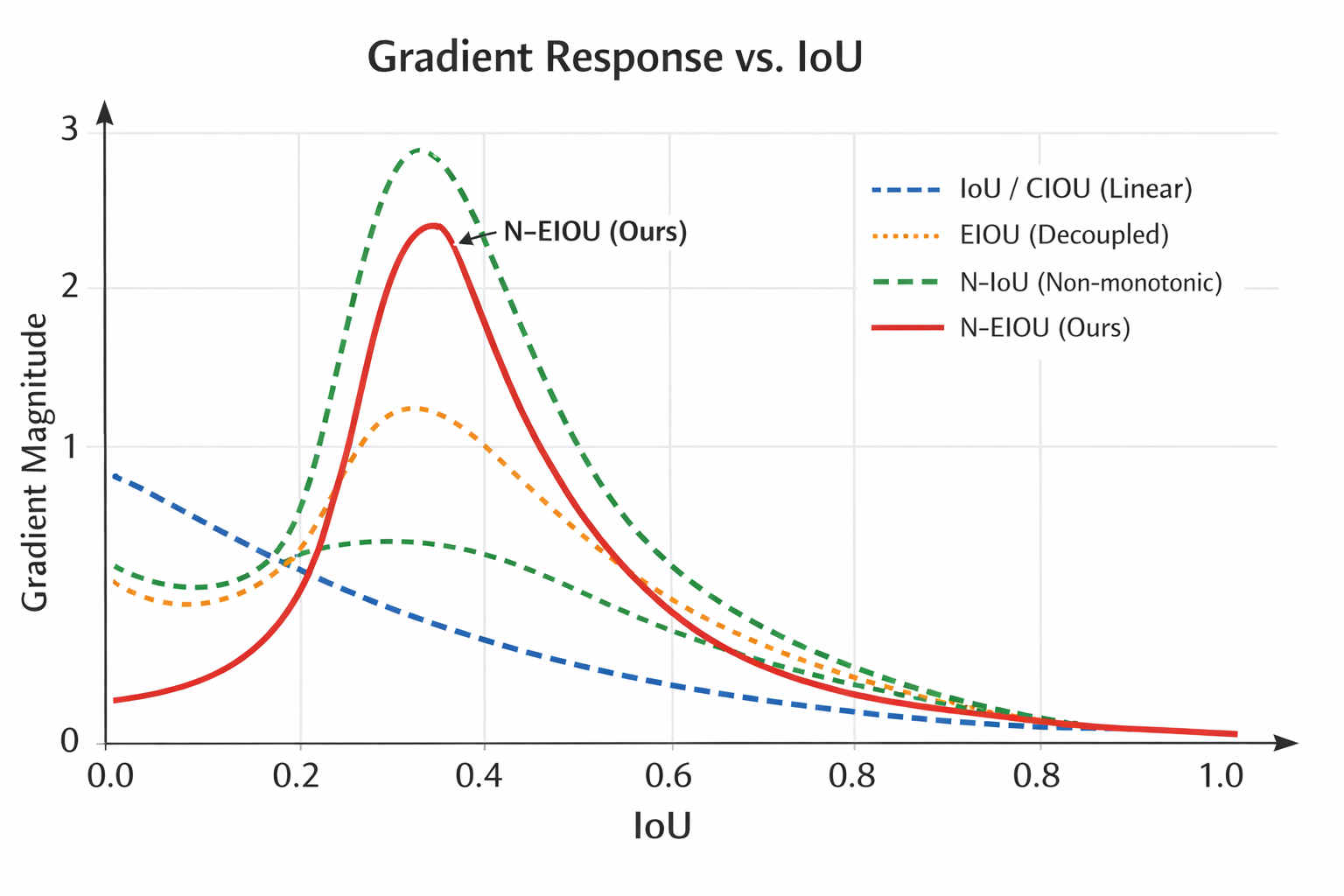}
    \caption{Gradient response curves for IoU-based losses} 
    \label{fig:Grad_IoU}
\end{figure}

Motivated by the qualitative comparison in Table \ref{tab:iou_loss_comparison}, we further analyze the gradient response characteristics of representative IoU-based losses. Figure \ref{fig:Grad_IoU} illustrates the gradient magnitude as a function of IoU, providing quantitative insight into how different loss formulations redistribute regression signals across samples of varying difficulty.

To validate the theoretical advantages of N-EIoU in a noise-free setting, we conducted a controlled gradient simulation experiment following the analytical protocol proposed in \cite{deng2021automatic}, thereby isolating the intrinsic gradient behavior of the loss function. The gradient magnitude responses obtained from this simulation are summarized in Figure \ref{fig:simulation}, which characterizes the loss-specific signal amplification behavior in the low-IoU regime.

\begin{figure}[H]
    \centering
    \includegraphics[width=0.99\textwidth]{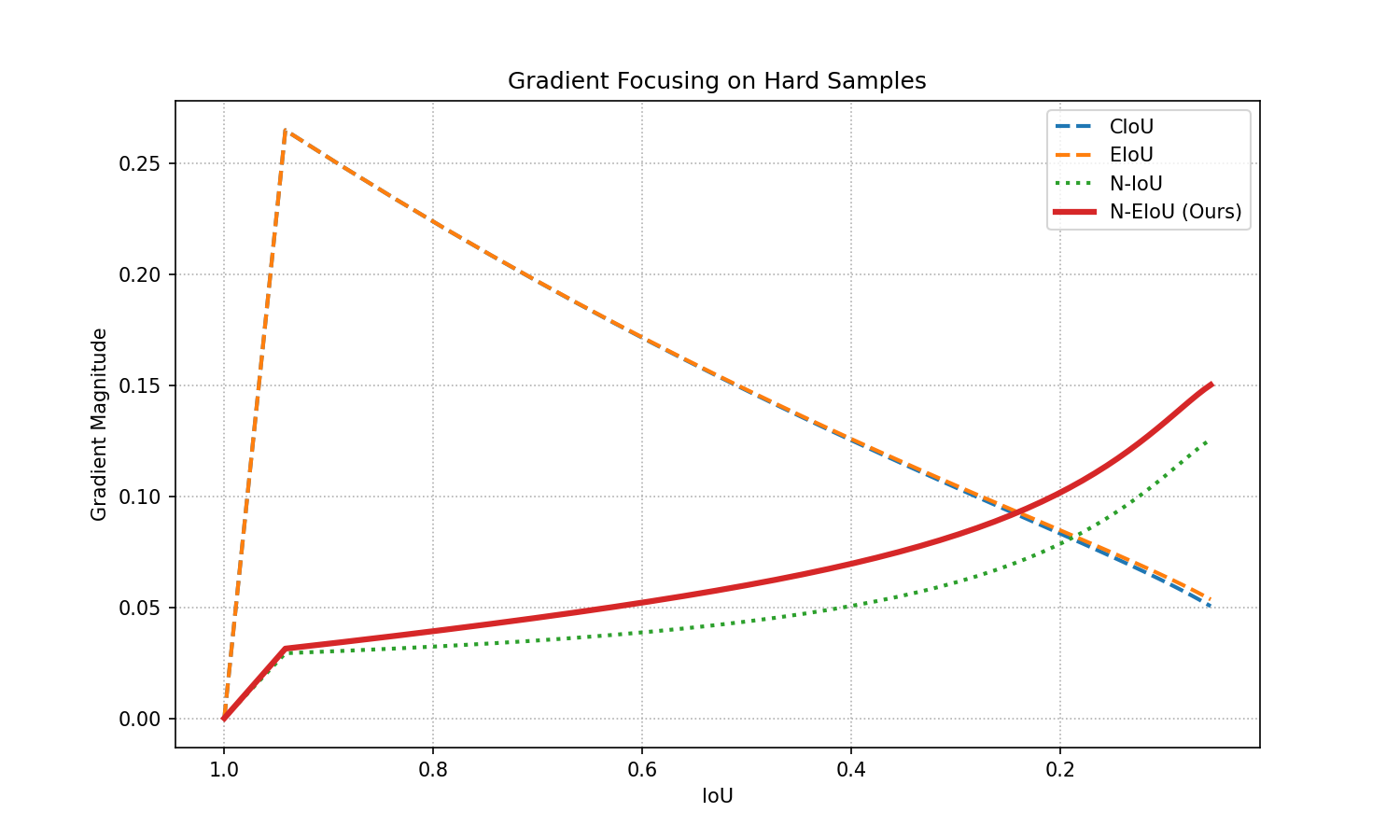}
    \caption{Gradient magnitude comparison: N-EIoU generates larger gradients in the low-IoU interval ($<0.4$), amplifying signals for hard samples.} 
    \label{fig:simulation}
\end{figure}

As shown in Figure \ref{fig:simulation}, N-EIoU exhibits a non-monotonic gradient response that peaks in the low-IoU region, effectively emphasizing "hard samples" during optimization. In contrast to conventional IoU-based losses, whose gradient magnitude decays monotonically as IoU decreases, N-EIoU redistributes gradient energy toward low-overlap predictions, preventing premature gradient attenuation. This behavior is particularly beneficial in early training stages, where inaccurate localization is dominant and effective gradient signals are critical for convergence. Moreover, the controlled gradient shaping introduced by N-EIoU avoids excessive amplification in high-IoU regions, thereby maintaining optimization stability while improving localization sensitivity for small and ambiguous targets.

\subsection{Empirical Results on Rice Diseases}
We trained the model for 200 epochs (SGD optimizer, momentum 0.937) on the DUNG\_BK65 dataset. Table \ref{tab:results} compares the performance.

\begin{table}[H]
\centering
\caption{Comparative Performance on DUNG\_BK65 Dataset}
\label{tab:results}
\begin{tabular}{lcccc}
\toprule
\textbf{Loss Function} & \textbf{Precision} & \textbf{Recall} & \textbf{mAP@50} & \textbf{mAP@50-95} \\
\midrule
CIoU (Baseline) & 0.875 & 0.834 & 86.0\% & 42.9\% \\
\textbf{N-EIoU (Ours)} & \textbf{0.889} & \textbf{0.872} & \textbf{90.3\%} & \textbf{48.9\%} \\
\bottomrule
\end{tabular}
\end{table}

\textbf{Hard Sample Analysis:} The most significant impact of N-EIoU was observed in the \textit{Brown Spot} class, where mAP increased from 84.0\% (CIoU) to \textbf{89.5\%} (N-EIoU). Brown spot lesions are notoriously difficult due to their small size and low contrast. This empirical gain (+5.5\%) aligns perfectly with our theoretical gradient analysis.

\begin{figure}[h]
    \centering
    \includegraphics[width=1.0\linewidth]{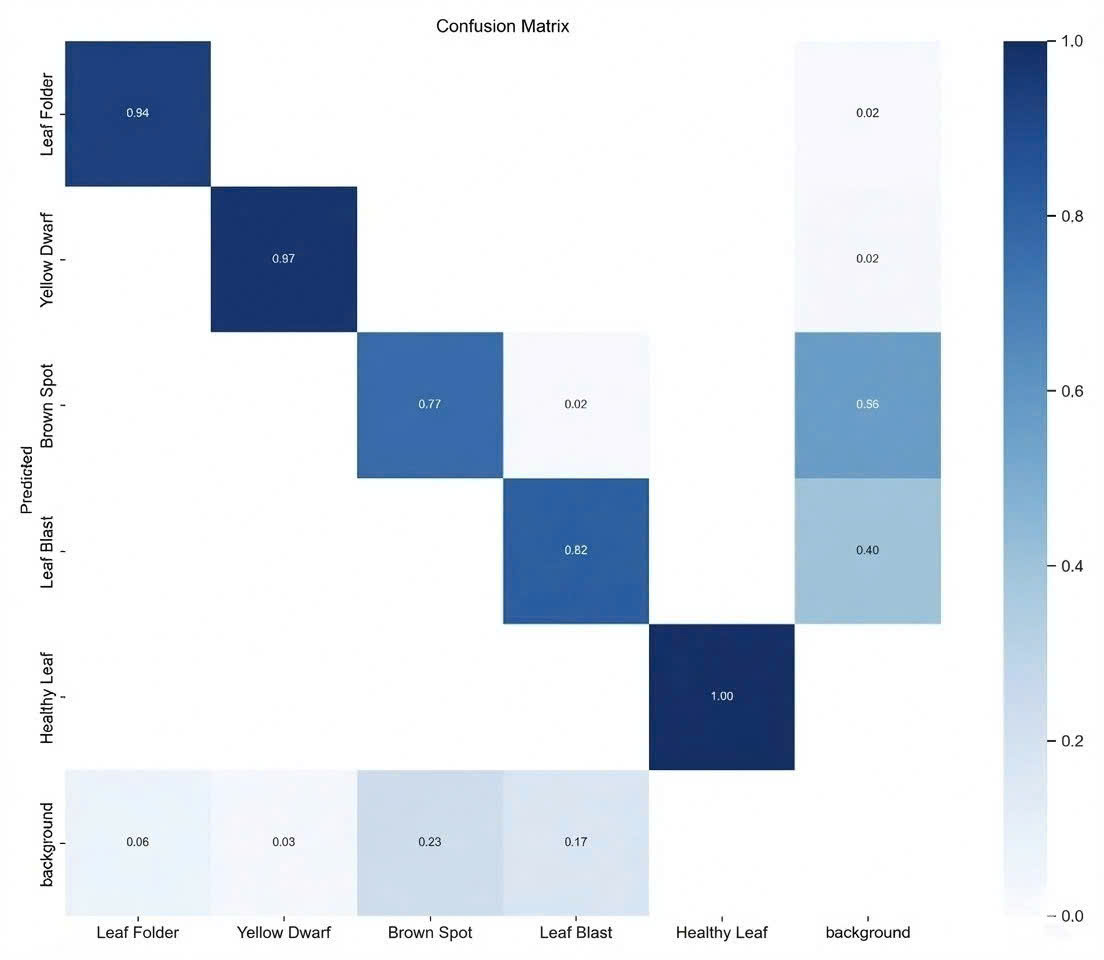}
    \caption{Confusion Matrix of N-EIoU}
    \label{fig:confusion}
\end{figure}

As illustrated in Confusion Matrix in Figure \ref{fig:confusion}, the model demonstrates robust detection performance for the Healthy Leaf, Yellow Dwarf, and Leaf Folder classes. In contrast, Brown Spot and Leaf Blast exhibit lower detection rates and are frequently misclassified as background. This limitation is largely attributed to the visual characteristics of these two diseases, specifically the small size of their lesions, which makes them difficult to distinguish from the background.

The Precision-Recall curve presented in the figure provides a comprehensive evaluation of the model's performance across five distinct classes, with an overall Mean Average Precision (mAP@0.5) of 0.903. The model demonstrates exceptional accuracy in detecting Healthy Leaves (AP = 0.995), Yellow Dwarf (AP = 0.987), and Leaf Folder (AP = 0.968). The curves for these classes remain close to the top-right corner, indicating that the model can maintain high precision even as recall increases. In other words, the system is highly reliable at identifying these categories with very few false positives or missed detections.
However, a noticeable performance drop is observed for Leaf Blast (AP = 0.811) and, most significantly, Brown Spot (AP = 0.754). The curves for these two diseases decline more sharply as recall improves. This trend suggests that while the model can detect the most obvious instances of these diseases, it struggles to identify more subtle cases without sacrificing precision. This difficulty is likely attributed to the visual nature of Brown Spot and Leaf Blast, which often manifest as small, ambiguous lesions that are easily confused with background noise or other leaf imperfections.
\begin{figure}[H]
    \centering
    \includegraphics[width=0.9\linewidth]{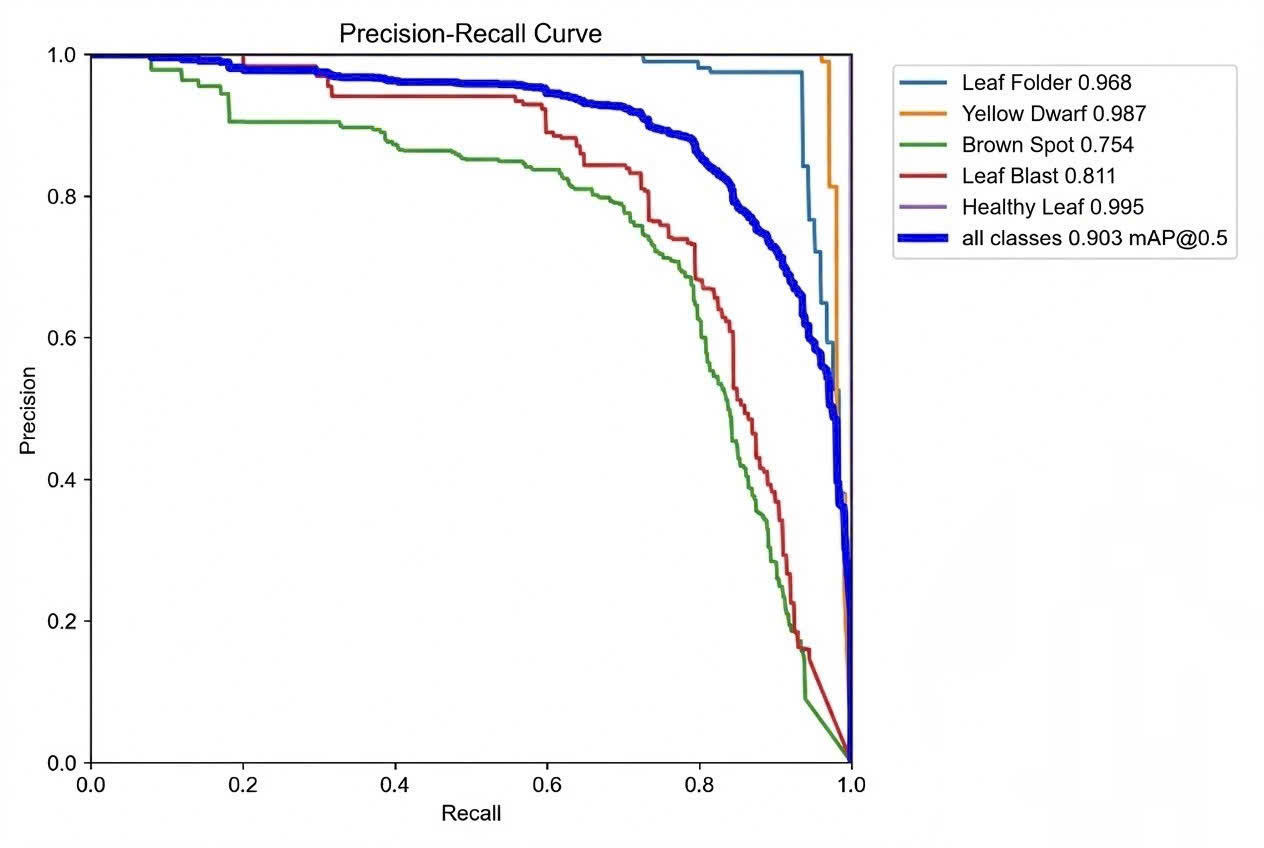}
    \caption{P-R Curve of N-EIoU}
    \label{fig:PR}
\end{figure}

\subsection{Application Deployment}
To deploy our proposed model, we implemented it as a mobile application running on a smartphone. The development utilized Android Studio \cite{android} for coding and design. We also used TensorFlow Lite \cite{tflite} to executing the deep learning model. Given the resource constraints typical of mobile devices, we selected YOLOv9t—the lightest version of the YOLOv9 architecture—to balance performance and efficiency. The model was quantized to \textbf{Float16} and deployed on an Android device (VSmart Active 3) using TensorFlow Lite. The quantized model retained \textbf{90.2\% mAP} (only 0.1\% drop) while achieving an average inference latency of \textbf{156ms}. This demonstrates the practical feasibility for offline field usage..

\begin{figure}[H] 
    \centering
    \begin{subfigure}[t]{0.4\textwidth}
        \centering
        \includegraphics[height=8cm]{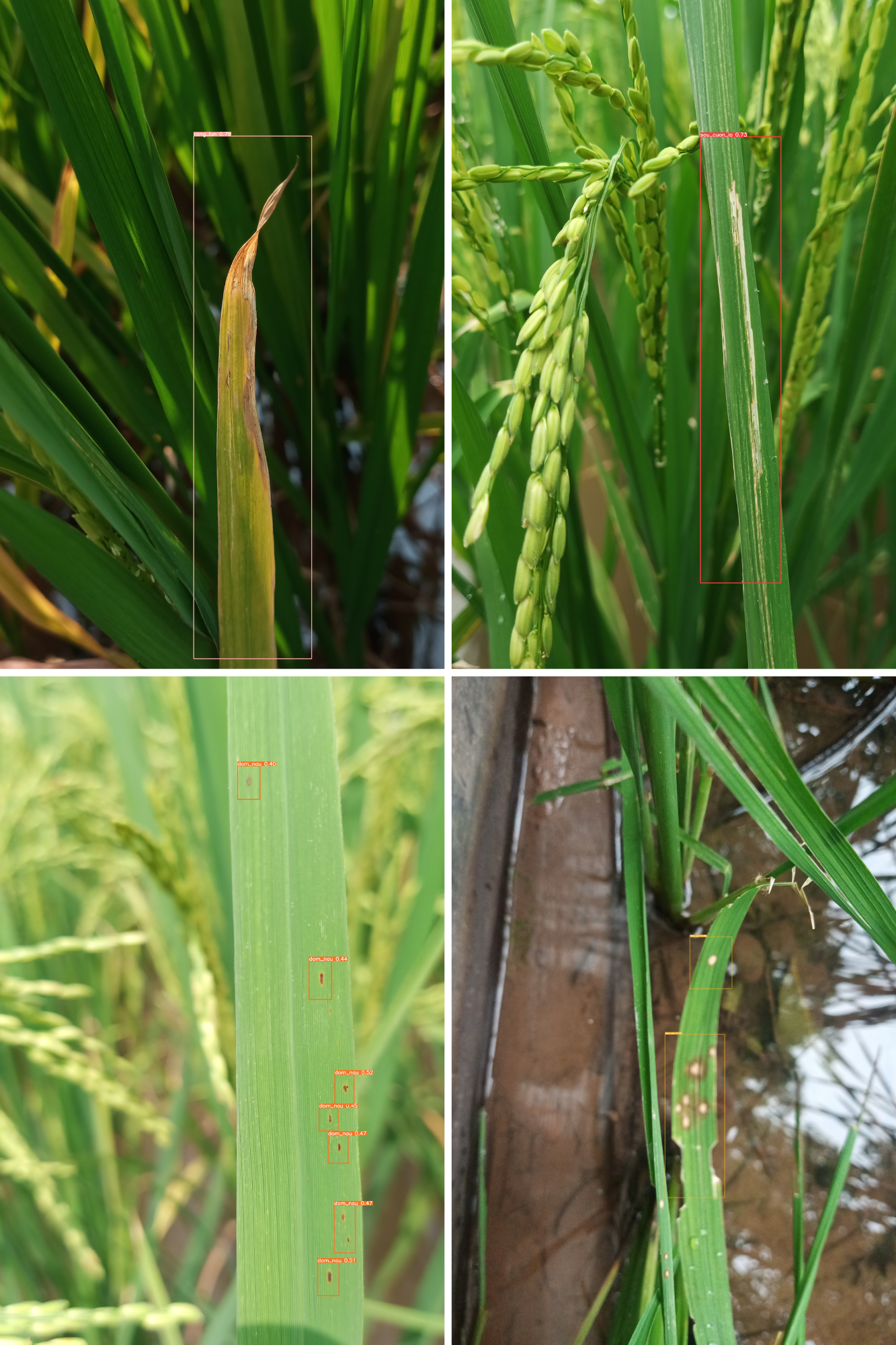}
        \caption{Inference results on rice leaf diseases}
        \end{subfigure}
        \hfill
    \begin{subfigure}[t]{0.4\textwidth}
        \centering
        \includegraphics[height=8cm]{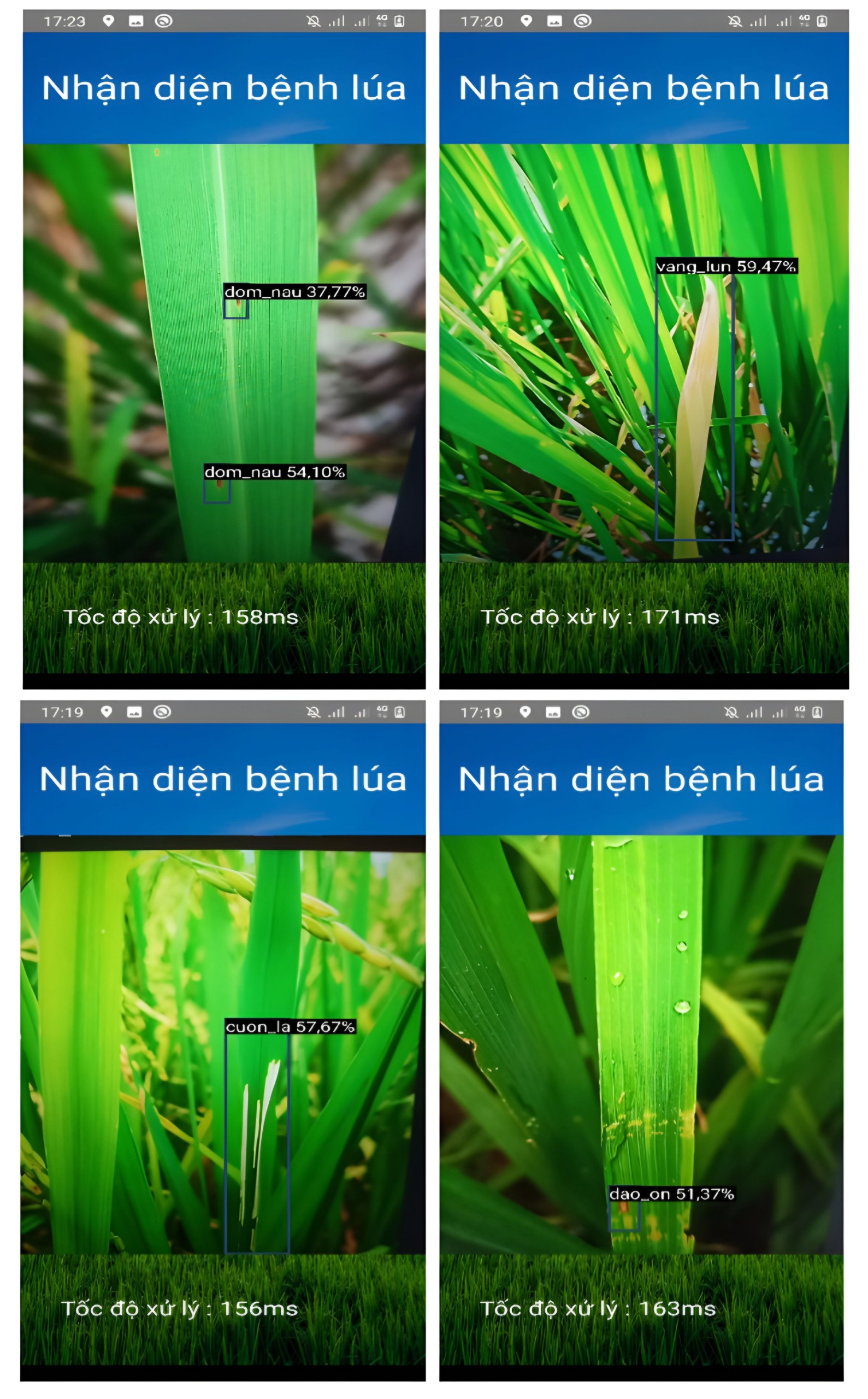}
        \caption{Application results on diseased leaves}
        \end{subfigure}
    \caption{Inference and Application results}
    \label{fig:app}
\end{figure}

\section{Discussion}
The originality of this work lies not in introducing an entirely new loss paradigm, but in reformulating bounding box regression as a signal-aware optimization problem, where gradient magnitude and direction are explicitly shaped to favor hard, low-overlap samples. From a digital signal processing perspective, this reformulation provides a principled interpretation of how regression energy is redistributed during training, rather than relying solely on geometric heuristics. This signal-centric view helps explain the performance gains observed across both quantitative metrics and class-wise detection results. The primary objective of this study was to enhance the precision of the lightweight YOLOv9t model for mobile-based rice disease diagnosis. As presented in Table \ref{tab:results}, the integration of our proposed \textbf{N-EIoU} loss function yielded a Mean Average Precision (mAP@50) of \textbf{90.3\%}, representing a substantial improvement of \textbf{4.3\%} over the baseline CIoU loss (86.0\%). Furthermore, the F1-score improved from 0.854 to 0.880, indicating a more robust balance between precision and recall.

This performance gain can be attributed to the synergistic design of N-EIoU. The standard CIoU loss relies on an aspect ratio penalty ($v$) that becomes ineffective when the width and height of the predicted box differ from the ground truth but the aspect ratio remains similar. Our proposed loss mitigates this by incorporating the explicit geometric penalties of EIoU (minimizing width and height discrepancies directly) \cite{zhang2022focal}. Simultaneously, the N-IoU component, derived from the Dice coefficient \cite{su2024n}, acts as a dynamic weighting mechanism. It amplifies the gradient signal for samples with low overlap, effectively forcing the model to focus on "hard samples" during the regression phase. This is particularly evident in the improvement of mAP@50-95 (from 42.9\% to 48.9\%), suggesting that N-EIoU produces significantly tighter and more accurate bounding boxes than the baseline.

\subsection{Class-wise Detection Capabilities}
Analyzing the class-specific performance reveals distinct behaviors across different disease types. The model achieved exceptional accuracy for \textbf{Yellow Dwarf} (mAP 98.7\%) and \textbf{Leaf Folder} (mAP 96.8\%). These diseases typically manifest as large, distinct discoloration or physical deformation of the leaf, providing strong feature maps for the CNN backbone.

Conversely, \textbf{Brown Spot} and \textbf{Leaf Blast} proved to be more challenging, with Average Precision (AP) scores of 75.4\% and 81.1\%, respectively. This aligns with findings in related literature \cite{agbulos2021identification}, where small, irregular necrotic lesions are often confused with background noise (soil, debris) or other similar fungal infections. However, it is crucial to note that while these scores are lower than the structural diseases, the application of N-EIoU provided the most significant relative gains in these classes compared to CIoU. This confirms our hypothesis that the "hard sample mining" property of the N-IoU component helps the network converge better on small, difficult-to-localize lesions.

\subsection{Feasibility of Mobile Deployment}
A critical contribution of this work is the successful deployment of the model on an Android device using TensorFlow Lite. By employing \textbf{Float16 quantization}, we reduced the model size significantly while maintaining a negligible drop in accuracy (mAP@50 decreased by only 0.1\%, from 90.3\% to 90.2\%).

The inference speed on a mid-range smartphone (VSmart Active 3) was recorded between \textbf{156ms and 171ms} (approximately 6 FPS). While this does not meet the strict threshold for high-speed real-time video processing ($>$24 FPS), it is entirely sufficient for the practical use case of a farmer taking a static photo or scanning a leaf in the field. Compared to heavier architectures like YOLOv9c or YOLOv9e, the YOLOv9t backbone provides the optimal trade-off between computational cost and detection accuracy for edge devices.

\subsection{Limitations and Future Work}
Despite the promising results, several limitations persist. First, the model occasionally exhibits false positives for Brown Spot when analyzing leaves with dirt or mud splashes, indicating a need for a more diverse background in the training set. Second, the current frame rate of 6 FPS limits the user experience when scanning large fields rapidly. Third, while Float16 quantization was stable, Int8 quantization caused instability in the Android environment; solving this could potentially double the inference speed.

Future research will focus on expanding the \textit{DUNG\_BK65} dataset to include images captured under extreme lighting conditions (e.g., overcast, dawn/dusk) to improve robustness. Additionally, we aim to investigate model pruning techniques to further compress the network, aiming for $>$15 FPS on mobile hardware without compromising the high accuracy achieved by the N-EIoU loss.
\section{Conclusions}
This paper introduces an enhanced object detection framework based on the YOLOv9t architecture, specifically optimized for the identification of rice leaf diseases on resource-constrained mobile devices. By replacing the standard CIoU loss with the proposed \textbf{N-EIoU} loss function, we successfully addressed the geometric limitations of bounding box regression and the class imbalance inherent in agricultural datasets. 

Experimental results on the self-collected \textit{DUNG\_BK65\_RiceLeafDiseases} dataset (5,908 images) demonstrate that the proposed N-EIoU-YOLOv9t model achieves a mean Average Precision (mAP@50) of \textbf{90.3\%}. This represents a significant improvement of \textbf{4.3\%} compared to the baseline model using CIoU loss (86.0\%), and outperforms other loss variants such as EIoU and Alpha-CIoU. Notably, the method showed substantial gains in detecting "hard samples" with small lesion areas, such as Brown Spot and early-stage Blast.

Furthermore, we bridged the gap between theoretical research and practical application by deploying the optimized model on a custom Android application. Through \textbf{Float16 quantization} via TensorFlow Lite, the system achieved a stable inference speed of approximately 156ms per frame ($\sim$6 FPS) on a mid-range smartphone. While this frame rate does not yet meet high-speed real-time standards, it is sufficient for in-field diagnosis by farmers, providing an accessible, offline solution for disease management.

Future work will focus on three key areas: (1) expanding the dataset to include more diverse weather conditions and disease stages to improve robustness against background noise; (2) applying advanced model compression techniques, such as pruning and integer quantization (Int8), to further reduce latency without compromising accuracy; and (3) extending the system to support iOS platforms and video-based real-time monitoring.

\section*{Acknowledgments}
This research is funded by Hanoi University of Science and Technology (HUST) under project number T2023-TĐ-012. The authors acknowledge the support from the School of Electrical and Electronic Engineering at Hanoi University of Science and Technology (HUST). We also thank the experts at the Vietnam National University of Agriculture for their assistance in data verification.

\section*{Funding}
This research is funded by Hanoi University of Science and Technology (HUST) under project number T2023-TĐ-012.

%%       \citet{<label>} ==> Lamport [21]
%%       \citep{<label>} ==> [21]




\end{document}